\documentclass[runningheads]{llncs}

\usepackage{eccv}

\usepackage{eccvabbrv}

\usepackage{graphicx}
\usepackage{booktabs}

\usepackage[ruled,linesnumbered, boxed, inoutnumbered]{algorithm2e}
\usepackage{multirow}
\usepackage[accsupp]{axessibility}  

\usepackage{hyperref}

\usepackage{orcidlink}

\begin{document}

\title{Deep Online Probability Aggregation Clustering} 


\author{Yuxuan Yan \and
Na Lu\thanks{Corresponding author}  \and
Ruofan Yan }

\authorrunning{Y. Yan, N. Lu et al.}

\institute{Systems Engineering Institute, Xi’an Jiaotong University \\
 \email{yan1611@stu.xjtu.edu.cn, lvna2009@mail.xjtu.edu.cn, yanruofan@stu.xjtu.edu.cn}\\ 
}

\maketitle

\begin{abstract}
  Combining machine clustering with deep models has shown remarkable superiority in deep clustering. It modifies the data processing pipeline into two alternating phases: feature clustering and model training. However, such alternating schedules may lead to instability and computational burden issues. To tackle these problems, we propose a centerless clustering algorithm called Probability Aggregation Clustering (PAC), enabling easy deployment in online deep clustering. PAC circumvents the cluster center and aligns the probability space and distribution space by formulating clustering as an optimization problem with a novel objective function. Based on the computation mechanism of the PAC, we propose a general online probability aggregation module to perform stable and flexible feature clustering over mini-batch data and further construct a deep visual clustering framework deep PAC (DPAC). Extensive experiments demonstrate that DPAC remarkably outperforms the state-of-the-art deep clustering methods.\footnote{The code is available at \url{https://github.com/aomandechenai/Deep-Probability-Aggregation-Clustering}}
\end{abstract}

\section{Introduction}
Clustering analysis \cite{arabie1996complexity} is a widely explored domain in the field of unsupervised learning, aiming to group the unlabeled samples into clusters that have common characteristics. Conventional machine clustering is favored by many researchers due to its significant interpretability and stable optimization. In recent years, deep clustering has received more attention due to its powerful representation extraction capabilities. Previous deep clustering models \cite{xie2016unsupervised, yang2017towards, guo2017improved, cai2022efficient} directly combine deep networks with machine clustering and utilize designed loss functions to guide both representation learning and clustering. For example, Deepcluster \cite{caron2018deep} and PCL \cite{li2020prototypical} decouple representation learning and clustering to leverage the offline pseudo labels of K-means (KM) to cluster images. Unfortunately, these offline methods typically require running multiple times of standard KM over the entire dataset, which brings much time and space complexity. Besides, simply grouping data in batches instead of the whole dataset to obtain online clustering causes collapsing and degradation issues. To address these problems, researchers have given two dominant solutions: batch clustering and contrastive clustering.

Batch clustering \cite{zhan2020online, deshmukh2022representation, jiao2022fine, nassar2023protocon} focuses on modifying the conventional machine clustering algorithms \cite{zhong2005efficient} to adapt the data flow of deep models, which has high extensibility. For example, Online Deep Clustering (ODC) \cite{zhan2020online} decomposes the standard KM process into batch clustering with memory banks and optimizes the clustering and network shoulder-to-shoulder (online) to facilitate stable learning. CoKe \cite{qian2022unsupervised} proposes the moving average strategy to reassignment pseudo labels and introduces Constrained K-means \cite{bradley2000constrained} into training to ensure the minimal size of clusters to avoid collapsing. Most existing batch clustering approaches focus more on center-based machine clustering algorithms, such as KM and fuzzy c-means (FCM) \cite{bezdek1984fcm}, which require specially designed center update rules. Moreover, center-based machine clustering is easily susceptible to the influence of cluster center \cite{franti2019much, arthur2007k}. Random initialization of cluster centers introduces instability to subsequent training. Partitioning based on nearest centers cannot provide fine-grained discrimination hyperplanes for clusters, affecting clustering performance. 

Recently, contrastive clustering \cite{shen2021you, li2022twin, tsai2021mice, znalezniak2023contrastive} has achieved significant success in online deep clustering. Contrastive methods perform online clustering by exploring multi-view correlations of data. Formally, instances are augmented into two views using random data augmentation to build contrastive frameworks. The clustering process is then performed by minimizing the designed contrastive loss. For example, PICA \cite{huang2020deep} proposes cluster-level contrastive loss based on contrastive framework to perform online deep clustering. However, the establishment of contrastive approaches needs a lot of artificial knowledge, including data augmentation, hyperparameter setting, and model architecture. Contrastive models often need thousands of epochs to reach convergence. Besides, they make a balanced assumption for clustering (i.e. each cluster has the same number of samples), which requires additional regular terms to constrain optimization and avoid crash problems (i.e. a few clusters have a majority of instances). The essence of contrastive clustering methods is to leverage the nearest-neighbor relationship of augmented instances in the semantic space to unsupervisedly train the classifier. Such semantic nearest-neighbor learning only uses a portion of data and its corresponding augmented version, failing to capture the global cluster relationship \cite{chen2020simple} and encode spatial embedding distribution.

In this work, considering the adverse effect of the cluster center, we first introduce a novel objective function quantifying the intra-cluster distances without cluster centers. Furthermore, inspired by fuzzy c-means, a concise optimization program is formulated by incorporating a fuzzy weighting exponent into an objective function. Then we build a centerless machine clustering algorithm called Probability Aggregation Clustering (PAC). In the optimization program of PAC, the probability of one sample belonging to a cluster is aggregated across samples with distance information in an iterative way. Unlike KM which assigns instances by cluster centers, PAC directly outputs probabilities which is more stable and easy to deploy in deep models. Therefore, we extend the PAC to the online probability aggregation module (OPA), a simple plug-in component for online deep clustering tasks. OPA seamlessly combines the calculation process of PAC with loss computation. It overcomes the disadvantages of both batch and contrastive clustering and implements efficient clustering. Besides, OPA does not impose any constraints on the size of clusters, mitigating the suboptimal solutions introduced by balanced clustering and obtaining more flexible partitioning. It computes clustering codes with the batches of data and updates the network by KL divergence, which leaves out the complicated clustering steps and trains the model in a supervised manner. Based on the above theories, a deep image clustering model Deep PAC (DPAC) is established, which ensures stable learning, global clustering, and superior performance. The major contributions of this work include:

\begin{itemize}
	\item A novel centerless partition clustering method PAC is proposed to implement clustering by exploring the potential relation between sample distribution and assignment probability. 
	\item An online deep clustering module OPA is exploited based on PAC, which encodes spatial distances into online clustering without incorporating plenty hyper-parameters and components. It leaves out the cluster size constraints to perform flexible partitioning.
	\item A simple end-to-end unsupervised deep clustering framework DPAC is established for stable and efficient clustering. DPAC achieves significant performance on five challenging image benchmarks compared with the state-of-the-art approaches.
\end{itemize}

\section{Related Work}
\subsubsection{Deep Clustering:}
Deep clustering methods \cite{chang2017deep, dang2021nearest, shen2021you} combine representation learning with clustering through deep models. ProPos \cite{huang2022learning} proposes the prototype scattering loss to make full use of K-means pseudo labels. Deepdpm \cite{ronen2022deepdpm} is a density-based approach, which does not require the preset number of class. Different from the above, recent deep clustering methods assume that the output is uniform. SwAV \cite{caron2020unsupervised} and SeLa \cite{asano2019self} adopt a balanced cluster discrimination task via the Sinkhorn-Knopp algorithm. SCAN \cite{van2020scan} leverages K-nearest-neighbor information to group samples. Its loss maximizes the agreements of assignments among neighbors, which inevitably need an additional balanced cluster constraint to avoid trivial solutions. SeCu \cite{qian2023stable} employs a global entropy constraint to relax the balanced constraint to a lower-bound size constraint that limits the minimal size of clusters.

\subsubsection{Machine Clustering:}
Machine clustering \cite{celebi2014partitional, van2003new, hornik2012spherical} tries to decompose the data into a set of disjoint clusters by machine learning algorithms. FCM \cite{bezdek1984fcm} obtains soft cluster assignment by alternately updating the fuzzy partition matrix and cluster center. Many modified7 methods \cite{lin2015cann, von2007tutorial,van2003new} aim at improving the performance and robustness of center-based clustering. In addition, nonparametric methods \cite{ester1996density, frey2007clustering} have received more and more attention in recent years. FINCH \cite{sarfraz2019efficient}  performs hierarchical agglomerative clustering based on first-neighbor relations without requiring a specific number of clusters. However, the complex clustering progresses involved in these algorithms hinder their easy deployment in neural networks.

\section{Method}
The following sections present the theoretical basis of our approach. We first derive a novel objective function and analyze how the proposed objective function relates to existing methods. Second, we present a scalable centerless clustering algorithm PAC. Finally, we extend PAC to a novel online clustering module OPA, and construct a novel online deep clustering model DPAC to learn the semantic knowledge of unlabeled data.

\subsection{Objective Function}\label{sec:Objective Function}
Let $\boldsymbol{X}=\lbrace \boldsymbol{x}_{1}, \boldsymbol{x}_{2}, \cdots , \boldsymbol{x}_{N} \rbrace$ be an $N$-point dataset, where $\boldsymbol{x}_{i}\in\mathbb{R}^{D\times 1}$ is the i-th $D$-dimensional instance. The clustering algorithm aims to divide $\boldsymbol{X}$ into $K$ mutually disjoint clusters, where $2\leq K<N$, $K \in \mathbb{N}$. $\boldsymbol{P}=[p_{i,k}]_{N\times K}$ is the soft partition matrix, $p_{i,k}$ is the probability of one sample belonging to certain cluster indicating the relationship between sample $\boldsymbol{x}_{i}$ and cluster $k$ which satisfies $	\boldsymbol{P}\in\{\Gamma^{N \times K}|\gamma_{i,k} \in [0,1],\forall i,k;\quad \sum_{k=1}^{K}\gamma_{i,k}=1, \forall i;\quad  0<\sum_{i=1}^{N}\gamma_{i,k}<N,\forall k\}$. And the cluster prediction of $ \boldsymbol{x}_{i}$ can be predicted by $\displaystyle \hat{p_{i}}=\arg\max\limits_{k}p_{i,k},\ 1<k\leq K$.

Different from the existing classical center-based methods \cite{bezdek1984fcm,wang2014optimized}, we utilize the inner product operation of probability vectors instead of cluster center to indicate cluster relations of samples. Formally, we multiply the inner product results with corresponding distance measurements to quantify the global intra-cluster distance of the data. The objective function $J_{pac}$ is defined as:
\begin{equation}\label{eq:eqn-1} 
		J_{pac} = \sum_{i=1}^{N} \sum_{j=1}^{N}\boldsymbol{p}_i^{\mathsf{T}}\boldsymbol{p}_j \Vert \boldsymbol{x}_i-\boldsymbol{x}_j \Vert^2,
\end{equation}   
where $\boldsymbol{p}_i=[p_{i,1}, p_{i,2}, \ldots,p_{i,K}]^{\mathsf{T}}$ is the probability vector. $\boldsymbol{p}_i^{\mathsf{T}}\boldsymbol{p}_j\in[0,1]$ can be regarded as the probability weight for $\Vert \boldsymbol{x}_i-\boldsymbol{x}_j \Vert^2$. By minimizing \cref{eq:eqn-1}, $\boldsymbol{p}_i^{\mathsf{T}}\boldsymbol{p}_j$ can be negatively related to $\Vert \boldsymbol{x}_i-\boldsymbol{x}_j \Vert^2$, which denotes the probabilities of instances consistent with nearby samples, but not with distant samples. 

\subsection{Relation to Existing Methods}
We provide a new perspective to further understand the proposed objective function. We summarize the difference between our method and Spectral Clustering (SC) \cite{von2007tutorial} and SCAN \cite{van2020scan}. The minimizing problem for $J_{pac}$ can be rewritten as:
\begin{equation}\label{eq:eqn-2} 
	\min\limits_{\boldsymbol{P} \in \Gamma^{N \times K}}{} Tr(\boldsymbol{P}^{\mathsf{T}} \widetilde{\boldsymbol{D}}_x \boldsymbol{P}),
\end{equation}  
where $\widetilde{\boldsymbol{D}}_x$ is the distances matrix, $\widetilde{d}_{i,j}=\Vert \boldsymbol{x}_i-\boldsymbol{x}_j \Vert^2$. Obviously, $ \widetilde{d}_{i,j}$ can be replaced by many other distance measurement. We use $L_2$ distance as the default distance measure in the following experiments. The graph partitioning problem of SC is formulated as:
\begin{equation}\label{eq:eqn-3} 
	\begin{aligned}
	&\min\limits_{\boldsymbol{H} \in \mathbb{R}^{N\times K}}{} Tr(\boldsymbol{H}^{\mathsf{T}} \widetilde{\boldsymbol{L}}_x \boldsymbol{H}),\\
	& \begin{array}{r}
		\mathrm{s.t.}\ \boldsymbol{H}^{\mathsf{T}}\boldsymbol{H}=\boldsymbol{I},\\
	\end{array} 
	\end{aligned}
\end{equation}  
where $\widetilde{\boldsymbol{L}}_x$ is the Laplacian matrix of graph. The indicator matrix $\boldsymbol{H}$ contains arbitrary real values with orthogonality constraint. The semantic clustering loss in SCAN can be reformulated as:
\begin{equation}\label{eq:eqn-4}
	\begin{aligned}
	&\max\limits_{\boldsymbol{P} \in \Gamma^{N\times K}}\sum_{i=1}^{N} \sum_{j \in \mathcal{N}_i} \log{\boldsymbol{p}_i^{\mathsf{T}}\boldsymbol{p}_j}-\lambda \mathcal{H}(\boldsymbol{P})\\
	\Leftrightarrow
	&\max\limits_{\boldsymbol{P} \in \Gamma^{N\times K}}Tr(\boldsymbol{P}^{\mathsf{T}} \widetilde{\boldsymbol{A}}_x \boldsymbol{P})-\lambda\mathcal{H}(\boldsymbol{P}),
	\end{aligned}
\end{equation}  
where $\mathcal{H}(\boldsymbol{P})=\sum_{k=1}^{K}\frac{\sum_{i=1}^{N}p_{i,k}}{N}\log \frac{\sum_{i=1}^{N}p_{i,k}}{N}$, $\mathcal{N}_i$ is the $K$ nearest neighbor set of instance $i$, $\widetilde{\boldsymbol{A}}_x$ is the adjacent matrix, $\widetilde{a}_{i,j}=1$ when $ j\in \mathcal{N}_i$, otherwise $\widetilde{a}_{i,j}=0$. $\lambda$ is the hyper-parameter. The second term  $\mathcal{H}(\boldsymbol{P})$ in \cref{eq:eqn-4} denotes balanced constrain of cluster. Compared with \cref{eq:eqn-3}, \cref{eq:eqn-2} transforms the partitioning problem in Euclidean space into the graph-cut problem. And different from balanced partitioning in \cref{eq:eqn-4}, we convert the maximum problem to the minimum problem to efficiently avoid trivial solutions. The intrinsical constraints of probability matrix $\boldsymbol{P}$ enable $J_{pac}$ directly clustering without using orthogonality and balanced constraints. Therefore, DPAC does not require additional clustering regular terms \cite{van2020scan, shen2021you, li2021contrastive} to avoid collapse and performs more flexible cluster assignment. Moreover, unlike only using neighbors to group, $J_{pac}$ introduces the distance information into optimization to obtain a global clustering. 

\subsection{Probability Aggregation Clustering}
\label{sec:Probability Aggregation Clustering}
The proposed \cref{eq:eqn-2} is a constrained optimization problem. Inspired by FCM, we incorporate the fuzzy weighting exponent $m$ into the objective function and obtain a scalable machine clustering algorithm based on the Lagrange method. The new objective function with $m$ can be formulated as:
\begin{equation}\label{eq:eqn-5} 
	\tilde{J}_{pac} =\sum_{i=1}^{N} \sum_{j=1}^{N}\varphi (i, j) \tilde{d}_{i,j},\quad \text{with } \varphi (i, j)=\sum_{k=1}^{K}p_{i,k}^mp_{j,k},
\end{equation} 
where $m\in(1,+\infty)$. The corresponding Lagrange function is:
\begin{equation}\label{eq:eqn-6}
		\tilde{L}_{pac}=\sum_{i=1}^{N} \sum_{j\neq i}\varphi (i, j) \tilde{d}_{i,j}+\sum_{i=1}^{N}\lambda_i(1-\sum_{k=1}^{K}p_{i,k})-\sum_{i=1}^{N}\sum_{k=1}^{K}\gamma_{i,k}p_{i,k},
\end{equation}  
where $\lambda_\cdot$ and $\gamma_{\cdot,\cdot}$ are the Lagrange multipliers respectively for the sum constraint and the non-negativity constraint on $\boldsymbol{P}$. The partial derivative of $\widetilde{L}_{pac}$ with respect to $p_{i,k}$ should be equal to zero at the minimum as:
 \begin{equation}\label{eq:eqn-7}
	\frac{\partial \tilde{L}_{pac}}{\partial p_{i,k}} = 2\sum_{j\neq i}mp_{i,k}^{m-1} p_{j,k} \tilde{d}_{i,j} - \lambda_i - \gamma_{i,k}=0. \\
\end{equation}
 And according to the Karush-Kuhn-Tucker conditions we have:
$1-\sum_{k=1}^{K}p_{i,k}=0,\  \gamma_{i,k}p_{i,k}=0,\  \gamma_{i,k} \geq 0,\ \forall i,k.$
For soft clustering, endpoints are generally unreachable during optimization. Therefore, we only consider the case when $p_{i,k}\in(0,1)$, $\gamma_{i,k}=0$. Let $ \alpha=1/(m-1)$, it can be obtained from \cref{eq:eqn-7} that
$p_{i,k} =\lambda_i^{\alpha}{(2m\sum_{j\neq i}p_{j,k} \tilde{d}_{i,j})}^{-\alpha}$. Considering the sum constraint, the equation becomes $\lambda_i^{\alpha}\sum_{k=1}^{K}{(2m\sum_{j\neq i}p_{j,k} \tilde{d}_{i,j})}^{-\alpha}=\sum_{k=1}^{K}p_{i,k}=1$. By solving  $\lambda_i$ and taking it into \cref{eq:eqn-7},  we can finally obtain:
\begin{equation}\label{eq:eqn-8}
	p_{i,k}=\frac{s_{i,k}^{-\alpha}}{\sum_{r=1}^{K}s_{i,r}^{-\alpha}},\quad
	\text{with } s_{i,k}=\sum_{j\neq i}p_{j,k}\tilde{d}_{i,j}.\\
\end{equation}

Take one element $p_{i,k}$ as a variable and all the rest elements as constant, $\boldsymbol{P}$ can be iteratively updated with \cref{eq:eqn-8}. $s_{i,k}$ aggregates the probabilities and distances to compute a score that $\boldsymbol{x}_i$ belongs to cluster $k$. In other words, PAC solves $p_{i,k}$ through all other instances instead of cluster centers. PAC only needs to initialize the $\boldsymbol{P}$ following approximately uniform distribution, that is $p_{i,k}\approx 1/K$. Therefore, PAC circumvents the delicate cluster center initialization problem caused by disparate data distributions in the feature space \cite{arthur2007k}. The detailed steps of PAC are summarized in \cref{alg:alg1}.

\begin{algorithm}[h]
	\footnotesize
	\caption{PAC Program}\label{alg:alg1}
	\KwIn{dataset $\boldsymbol{X}$; weighting exponent $m$; cluster number $K$; initialization $\boldsymbol{P}$.}
	\While{not converage}{\For{$i\leftarrow 1$ \KwTo $N$}{\For{$k\leftarrow 1$ \KwTo $K$}{$p_{i,k} \gets \text {\cref{eq:eqn-8}}$ }}
	}
	\KwOut {Clustering result $\boldsymbol{P}$}
\end{algorithm}

\subsection{Online Probability Aggregation}\label{sec:Online Probability Aggregation}
 A deep neural network $\hat{\boldsymbol{x}}_i=f (\boldsymbol{I}_i)$ maps data $\boldsymbol{I}_i$ to feature vector $\hat{\boldsymbol{x}}_i$. And a classifier $h$ maps $\boldsymbol{x}_i$ to K-dimensional class probability $\hat{\boldsymbol{p}}_i$. We proposed a novel online clustering module OPA, which combines the optimization process of PAC with loss computation to generate pseudo labels step by step. Specifically, $B$ is the size of the mini-batch in the current epoch, OPA has two alternate steps:
 \subsubsection{Target Computation:} \cref{sec:Probability Aggregation Clustering} demonstrates the optimization program for a single variable, we extend it to the matrix to adopt multivariable. Given the current model $h \circ f$, the clustering score $\boldsymbol{S}\in\mathbb{R}^{+B\times K}$ is calculated by:
\begin{equation}\label{eq:eqn-9}
	\boldsymbol{S}=\widetilde{\boldsymbol{D}}_{\hat x}\hat{\boldsymbol{P}}.
\end{equation} 
The target clustering code $\boldsymbol{Q}\in{\Gamma}^{B\times K}$ can be obtained by normalizing $\boldsymbol{S}$, $q_{i,k}={s_{i,k}^{-\alpha}}/{\sum_{r=1}^{K}s_{i,r}^{-\alpha}}$. We call the operation in \cref{eq:eqn-9} as online probability aggregation. The probability outputs form the classifier are aggregated by matrix multiplication to compute corresponding scores, which not only incorporates historical partitioning knowledge but also encodes distance information.

\subsubsection{Self-labeling:}
Given the current target clustering code $\boldsymbol{Q}$, the whole model $h \circ f$ is updated by minimizing the following KL divergence: 
\begin{equation}\label{eq:eqn-10}
	KL(\boldsymbol{Q}\parallel\hat{\boldsymbol{P}})=\sum_{i=1}^{N} \sum_{k=1}^{K}q_{i,k}\log \frac{q_{i,k}}{\hat{p}_{i,k}}
\end{equation} 
Different from directly leveraging $J_{pac}$ in \cref{eq:eqn-1} as clustering loss, OPA trains the model in a supervised way instead of solving the clustering problem in \cref{eq:eqn-2} exactly. The pseudo code of OPA is illustrated in \cref{alg:alg2}, which only involves a mini-batch matrix multiplication and power, so the computation cost of OPA equals general loss. 

\begin{algorithm}[h]
	\footnotesize
	\caption{\footnotesize Pseudo code for OPA in pytorch-style}\label{alg:alg2}
	\KwIn{distance matrix $D$; probability matrix $P$; weighting exponent $m$.}
	\qquad$S=$ torch.matmul$(D$.detach()$,P)$\tcp*[f]{Aggregate Probability}\\
	\qquad$S=$ torch.pow$(S,-1/(m-1))$\tcp*[f]{Scale Up}\\
	\qquad$Q=S/S$.sum(1).view(-1,1)\tcp*[f]{ Normalize to 1}\\
	\KwOut {$(Q*\log Q-Q*\log P)$.sum(1).mean()\tcp*[f]{KL divergence loss}}
\end{algorithm}

\subsection{Deep Probability Aggregation Clustering}
With the proposed loss function, we construct an online deep clustering framework DPAC, which has two heads: contrastive learning and online clustering. Let $\hat{\boldsymbol{I}}^1_i$ and $\hat{\boldsymbol{I}}^2_i$ denote two-view features of $\hat{\boldsymbol{I}}_i$ generated by random image augmentation. We reformulate the standard contrastive loss in SimCLR \cite{chen2020simple} as weight contrastive loss (WCL) to mitigate the semantic distortion caused by negative samples. The weight contrastive loss $\ell(\hat{\boldsymbol{X}}^1, \hat{\boldsymbol{X}}^2, \hat{\boldsymbol{P}})$ is defined as:
\begin{equation}\label{eq:eqn-11}
	\begin{split}
		-\sum_{i=1}^{N}\log{\frac{\exp{(\hat{\boldsymbol{z}}_i^{1\mathsf{T}}\hat{\boldsymbol{z}}_i^2/\tau)}}{\sum_{j\neq i}\hat{w}_{i,j}\exp{(\hat{\boldsymbol{z}}_i^{1\mathsf{T}}\hat{\boldsymbol{z}}_j^1/\tau)}+\sum_{j=1}^N\hat{w}_{i,j}\exp{(\hat{\boldsymbol{z}}_i^{1\mathsf{T}}\hat{\boldsymbol{z}}_j^2/\tau)}}},
	\end{split}
\end{equation}
where $\tau$ is the temperature hyper-parameter, $\hat{\boldsymbol{z}}_i$ is the normalized feature projected by projector $g$, where $\hat{\boldsymbol{z}}_i=g(\hat{\boldsymbol{x}}_i)/\Vert g(\hat{\boldsymbol{x}}_i)\Vert$. $\hat{w}_{i,j}=(1-\hat{\boldsymbol{p}}_i^{\mathsf{T}}\hat{\boldsymbol{p}}_j)$ is a gate coefficient, which filters the negative samples that belong to same cluster as $\hat{\boldsymbol{x}}_i$.

In pre-training step, due to the lack of cluster information, $\hat{\boldsymbol{P}}$ is set to the uniform, $\hat{p}_{i,j}=1/K$, $\forall i,j$. And DPAC is pre-trained by the pairwise contrastive loss: $\frac{1}{2}[\ell(\hat{\boldsymbol{X}}^1, \hat{\boldsymbol{X}}^2, \hat{\boldsymbol{P}})+\ell(\hat{\boldsymbol{X}}^2, \hat{\boldsymbol{X}}^1, \hat{\boldsymbol{P}})]$. Then in clustering step, the whole model is updated by minimizing the sum of contrastive and clustering loss:
\begin{equation}\label{eq:eqn-12}
	\min\limits_{\boldsymbol{\theta}_{f,g,h}}	\frac{1}{2}[\ell(\hat{\boldsymbol{X}}^1, \hat{\boldsymbol{X}}^2, \hat{\boldsymbol{P}})+\ell(\hat{\boldsymbol{X}}^2, \hat{\boldsymbol{X}}^1, \hat{\boldsymbol{P}})]+KL(\boldsymbol{Q}\parallel\hat{\boldsymbol{P}}^1),
\end{equation} 
\begin{equation}\label{eq:eqn-13}
	\min\limits_{\boldsymbol{\theta}_{f,g,h}}	\frac{1}{2}[\ell(\hat{\boldsymbol{X}}^1, \hat{\boldsymbol{X}}^2, \hat{\boldsymbol{P}})+\ell(\hat{\boldsymbol{X}}^2, \hat{\boldsymbol{X}}^1, \hat{\boldsymbol{P}})]+\frac{1}{N} Tr(\hat{\boldsymbol{P}}^{1\mathsf{T}} \widetilde{\boldsymbol{D}}_{\hat{x}} \hat{\boldsymbol{P}}^1),
\end{equation} 
where $\boldsymbol{\theta}_{f,g,h}$ are the parameters of the neural network, classifier, and projector, respectively. \cref{eq:eqn-12} is the deep clustering method based on OPA mentioned in \cref{sec:Online Probability Aggregation}. \cref{eq:eqn-13} is the deep clustering method that directly minimizes $J_{pac}$ in \cref{sec:Objective Function}. The overall training procedure is shown in \cref{alg:alg3}. Moreover, for fair comparison in subsequent experiments, we also implement a self-labeling fine-tuning operation as \cite{li2022twin, sohn2020fixmatch} to further improve the clustering performance.

\begin{algorithm}[h]
	\footnotesize
	\caption{\footnotesize Training algorithm for DPAC}\label{alg:alg3}
	\KwIn{image set $\boldsymbol{I}$; clustering epochs $E$; batch size $B$; weighting exponent $m$.}
	\For{$epoch\leftarrow 1$ \KwTo $E$}{
		Sample a mini-batch $\{\boldsymbol{I}_i\}_{i=1}^B$ and
		conduct augmentations$\{\boldsymbol{I}_i^1, \boldsymbol{I}_i^2\}_{i=1}^B$;\\
		Get $\{\hat{\boldsymbol{x}}_i,\hat{\boldsymbol{x}}_i^1,\hat{\boldsymbol{x}}_i^2,\hat{\boldsymbol{p}}_i,\hat{\boldsymbol{p}}_i^1\}_{i=1}^B$ through forward propagation;\\
		\If{choose OPA as optimal object}{Compute clustering codes $\{\boldsymbol{q}_i\}_{i=1}^B$ by \cref{alg:alg2} with $\{\hat{\boldsymbol{x}}_i,\hat{\boldsymbol{p}}_i\}_{i=1}^B$;\\	Compute overall loss $\mathcal{L}$ by \cref{eq:eqn-12} with $\{\hat{\boldsymbol{x}}_i^1,\hat{\boldsymbol{x}}_i^2,\hat{\boldsymbol{p}}_i,\hat{\boldsymbol{p}}_i^1,\boldsymbol{q}_i\}_{i=1}^B$ \;}
		\If{choose $J_{pac}$ as optimal object}{Compute overall loss $\mathcal{L}$ by \cref{eq:eqn-13} with $\{\hat{\boldsymbol{x}}_i^1,\hat{\boldsymbol{x}}_i^2,\hat{\boldsymbol{p}}_i,\hat{\boldsymbol{p}}_i^1\}_{i=1}^B$\;}
		Update $\boldsymbol{\theta}_f$, $\boldsymbol{\theta}_g$, $\boldsymbol{\theta}_h$ through gradient descent to minimize $\mathcal{L}$;
	}
	\KwOut {Deep clustering model $h\circ f$}
\end{algorithm}

\section{Experiment}
\subsubsection{Dataset:}Four real-world datasets and five widely used natural image datasets are involved to evaluate the clustering ability of PAC and DPAC. The details of the datasets are summarized in the \cref{tab:tab1}. For CIFAR-100, we used its 20 super-classes rather than 100 classes as the ground truth. For STL-10, its 100,000 unlabeled images are additionally used in the pre-training
step of DPAC. ImageNet-10 and ImageNet-Dogs are subsets of ImageNet-1k.  Clustering accuracy (ACC), normalized mutual information (NMI), and adjusted random index
(ARI) are adopted to compare the clustering results.
\begin{table}[t]
	\scriptsize
	\centering
	\caption{Dataset settings for our experiments.}
	\label{tab:tab1}
	\begin{tabular}{c|ccc|c|ccc}	
		\toprule 
		Dataset   & Sample &Class  &Size &  Dataset  & Sample &Class  &Dimension\\
		\midrule
		CIFAR-10 \cite{krizhevsky2009learning}  & 60,000 &10  &32$\times$32 & Coil-100 \cite{nene1996columbia} & 7,200 & 100 & 49,152\\
		CIFAR-100 \cite{krizhevsky2009learning}  & 60,000 &20  &32$\times$32 & Isolet \cite{cole1990isolet} & 7,796 & 26 & 617 \\
		STL-10 \cite{coates2011analysis}  & 13,000 &10  &224$\times$224  &Pendigits \cite{alimoglu1997combining} & 10,992 & 10 & 16\\
		ImageNet-10 \cite{chang2017deep} & 13,000 &10  &224$\times$224 &MNIST \cite{deng2012mnist} &10,000 & 10 & 784\\
		ImageNet-Dogs \cite{chang2017deep} & 19,500 &15  &224$\times$224 &&&&  \\
		\bottomrule
	\end{tabular}
\end{table}

\subsection{Probability Aggregation Clustering}
\subsubsection{Hyperparameter and Method Setting}
The effectiveness of the proposed PAC is verified by comparing it with multiple clustering methods on nine datasets. The $m$ of PAC is set to 1.03 for all datasets. The threshold value of RCC \cite{shah2017robust} is set to 1. The weighting exponent $m$ of FCM is set to 1.1 for real-world datasets and 1.05 for natural image datasets. We predefine $K$ for all algorithms except FINCH \cite{sarfraz2019efficient}. All algorithms are initialized randomly and run 10 times. The mean and variance of 10 times run are taken as comparison results.

\subsubsection{Algorithm Scalability}
The clustering results of the real-world datasets, which consist of samples with varying numbers, classes, and dimensions, are summarized in \cref{tab:tab2}. PAM and RCC time out due to the high dimensionality of Coil-100. PAC outperforms all the compared clustering algorithms on Coil-100 and Isolet but is not as effective as RCC on Mnist and Pendigit, which is specially designed for entangled data. The robustness and performance of PAC surpass center-based methods by a large margin.
Moreover, we also provide the clustering results on neural network feature data in \cref{tab:tab3} to explore the ability of PAC to handle data extracted by neural networks. RCC experience extreme performance degradation on neural network extracted data, so we exclude it from the comparison. PAC also performs well in processing neural network data. The improvement is not significant in CIFAR-100 and ImageNet-Dogs. One possible explanation is that these datasets give subtle differences in object classes, causing the pretrained representations to be indistinguishable.

\begin{table}[t]
	\caption{Clustering results (Avg$\pm$Std) and average time (s) of PAC on real-world datasets.  The best and second-best results are shown in bold and underlined, respectively. Metric: ACC (\%).}
	\label{tab:tab2}
	\centering
	\scriptsize
	\begin{tabular}{l| c| c| c| c | c |c| c| c}
		\toprule
		\textbf{Method}& Coil-100& Isolet & Pendigits & MNIST &\multicolumn{4}{c}{Average Time}  \\
		\midrule
		KM \cite{wang2014optimized} & 56.4$\pm$1.7 &52.7$\pm$4.5 &67.0$\pm$4.7	&53.0$\pm$3.6 &98.1 &0.2&0.05&0.07\\
		PAM \cite{van2003new} & N/A &55.5$\pm$0.0	&75.6$\pm$2.5	&47.2$\pm$1.7 &N/A&341.9&141.6&124.0\\
		FCM \cite{bezdek1984fcm} & 61.6$\pm$1.2	&\underline{55.8}$\pm$2.3	&70.5$\pm$2.1	&56.6$\pm$2.6 &2001.5&8.6 &0.9&0.6\\
		SC \cite{von2007tutorial} &58.2$\pm$0.7	&53.5$\pm$2.5	&62.4$\pm$4.2	&54.6$\pm$2.2 &11.7& 3.4& 5.8& 6.2\\
		SPKF \cite{hornik2012spherical} &59.7$\pm$1.3	&55.2$\pm$2.0	&71.4$\pm$4.4	&53.9$\pm$2.7 &101.6&0.6&0.07&0.2\\
		RCC \cite{shah2017robust}  &N/A	&15.3$\pm$0.0	&\textbf{79.6}$\pm$0.0	&\textbf{65.7}$\pm$0.0 &N/A&122.8&6.9&6.9\\
		FINCH \cite{sarfraz2019efficient} &56.4$\pm$0.0	&47.5$\pm$0.0	&62.7$\pm$0.0	&57.9$\pm$0.0 &15.1&0.5&0.05&0.05\\
		\midrule
		PAC &\textbf{65.1}$\pm$1.5	&\textbf{61.8}$\pm$0.0	&\underline{78.0}$\pm$0.0	&\underline{59.7}$\pm$3.6 &5179.0&249.6&153.6&423.4\\
		\bottomrule
	\end{tabular}
\end{table}

\begin{table}[t]
	\caption{Clustering results (Avg$\pm$Std) of PAC on deep features. Metric: ACC (\%).}
	\label{tab:tab3}
	\centering
	\scriptsize
	\begin{tabular}{l| c| c| c| c| c}
		\toprule
		\textbf{Method}& CIFAR-10& CIFAR-100 & STL-10 & ImageNet-10 & ImageNet-Dogs\\
		\midrule
		KM \cite{wang2014optimized} & 76.8$\pm$6.8 &41.8$\pm$1.7 &\underline{66.8}$\pm$4.3	&76.8$\pm$6.8 &41.8$\pm$1.7\\
		PAM \cite{van2003new} & 77.8$\pm$2.5 &41.0$\pm$1.1	&64.3$\pm$4.8	&79.9$\pm$4.6 &\textbf{52.6}$\pm$3.1\\
		FCM \cite{bezdek1984fcm} & 75.9$\pm$2.1	&42.3$\pm$0.7	&66.6$\pm$4.7	&75.9$\pm$2.1 &42.3$\pm$0.7\\
		SC \cite{von2007tutorial} &83.5$\pm$0.0	&40.0$\pm$1.1	&63.8$\pm$2.9	&\underline{82.9}$\pm$1.3 & 47.6$\pm$1.4\\
		SPKF \cite{hornik2012spherical} &75.9$\pm$5.7	&\underline{42.9}$\pm$1.9	&65.8$\pm$5.5	&80.6$\pm$7.6 & \underline{49.1}$\pm$3.8\\
		FINCH \cite{sarfraz2019efficient} &49.2$\pm$0.0	&32.0$\pm$0.0	&42.9$\pm$0.0	&52.6$\pm$0.0 &43.8$\pm$0.0\\
		\midrule
		PAC &\textbf{87.1}$\pm$0.0	&\textbf{43.8}$\pm$0.7	&\textbf{74.9}$\pm$2.6	&\textbf{95.8}$\pm$0.0 &47.3$\pm$3.9\\
		\bottomrule
	\end{tabular}
\end{table}

\subsubsection{Parameter Sensibility Analysis}
\begin{figure}[h]
	\centering
	\includegraphics[width=0.51\linewidth]{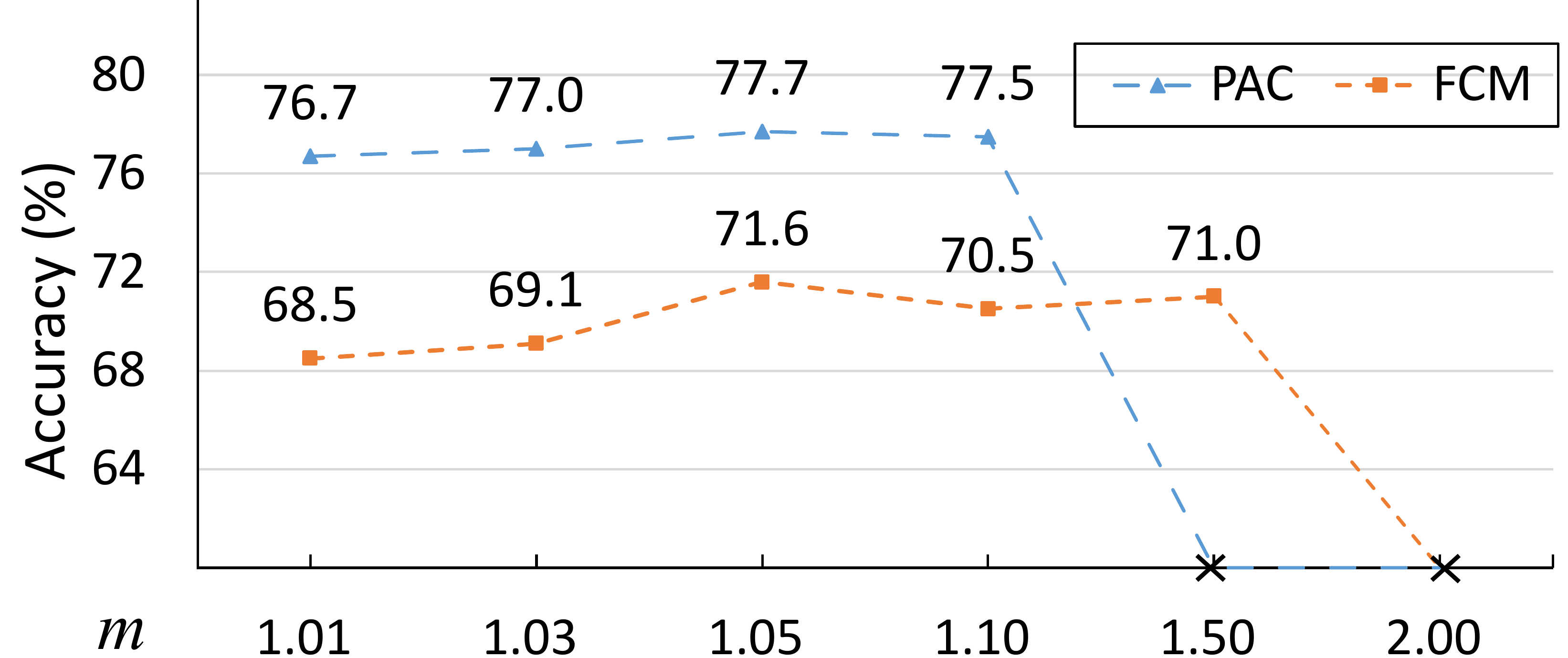}
	\caption{The effect of weighting exponent $m$ in PAC and FCM.}
	\label{fig:fig1}
\end{figure}

We evaluate the parameter sensitivity of $m$ for both FCM and PAC on Pendigits. \cref{fig:fig1} reports the average ACC  for different $m$. It was indicated that in comparison to FCM, PAC has a narrower optimal range of $m$ and smaller results variance, which is not sensitive to parameter $m$. 

\subsubsection{Time Complexity Analysis}
The average calculation time for each algorithm is listed in \cref{tab:tab2}. The computational complexity of PAC is analyzed in this section. It takes $\mathcal{O}(N)$ time to calculate $\sum_{j\neq i}p_{j,k}\tilde{d}_{i,j}$ in \cref{eq:eqn-8}. And PAC updates entire $\boldsymbol{P}$ by $NK$ iterations. So the time complexity PAC is $\mathcal{O}(N^2K)$, which is the square complexity.

\subsection{Deep Probability Aggregation Clustering}
\begin{table}[t]
	\caption{Performance comparison of deep clustering methods on five benchmarks.  The best and second-best results are shown in bold and underlined, respectively. Metrics: NMI / ACC / ARI (\%). Temu$^*$ incorporates extra ImageNet-1k data to pretrain the model, so we exclude it in comparison. $_1$ denotes online deep clustering methods, while $_2$ denotes offline deep clustering methods. Cluster const. denotes cluster size constraint.}
	\label{tab:tab4}
	\centering
	\resizebox{\linewidth}{!}{
		\begin{tabular}{l|c| ccc| ccc| ccc| ccc| ccc}
			\toprule
			\multirow{2.5}{*}{\textbf{Method}}
			&Cluster
			&\multicolumn{3}{c}{\textbf{CIFAR-10}} 
			&\multicolumn{3}{c}{\textbf{CIFAR-100}}
			&\multicolumn{3}{c}{\textbf{STL-10}}
			&\multicolumn{3}{c}{\textbf{ImageNet-10}}
			&\multicolumn{3}{c}{\textbf{ImageNet-Dogs}}\\
			\cmidrule(r){3-5}
			\cmidrule(r){6-8}
			\cmidrule(r){9-11}
			\cmidrule(r){12-14}
			\cmidrule(r){15-17}
			&const.  &NMI&ACC&ARI &NMI&ACC&ARI  &NMI&ACC&ARI &NMI&ACC&ARI &NMI&ACC&ARI\\
			\midrule
			PICA$_1$ \cite{huang2020deep} & \checkmark   &{59.1}&{69.6}&{51.2} 
			&{31.0}&{33.7}&{17.1} 
			&{61.1}&{71.3}&{53.1}
			&{80.2}&{87.0}&{76.1}
			&{35.2}&{35.2}&{20.1}\\	  
			PCL$_2$ \cite{li2020prototypical} &   &{80.2}&{87.4}&{76.6} 
			&{52.8}&{52.6}&{36.3} 
			&{41.0}&{71.8}&{67.0}
			&{84.1}&{90.7}&{82.2}
			&{44.0}&{41.2}&{29.9}\\
			IDFD$_2$ \cite{tao2021clustering} &  &{71.1}&{81.5}&{66.3} 
			&{42.6}&{42.5}&{26.4} 
			&{64.3}&{75.6}&{57.5}
			&\underline{89.8}&\underline{95.4}&\underline{90.1}
			&{54.6}&{59.1}&{41.3}\\
			NNM$_1$ \cite{dang2021nearest} & \checkmark  &{74.8}&{84.3}&{70.9} 
			&{48.4}&{47.7}&{31.6} 
			&{69.4}&{80.8}&{65.0}
			&{-}&{-}&{-}
			&{-}&{-}&{-}\\	
			CC$_1$ \cite{li2021contrastive} & \checkmark  &{70.5}&{79.0}&{63.7} 
			&{43.1}&{42.9}&{26.6} 
			&{76.4}&{85.0}&{72.6}
			&{85.9}&{89.3}&{82.2}
			&{44.5}&{42.9}&{27.4}\\	
			GCC$_1$ \cite{zhong2021graph}  & \checkmark  &{76.4}&{85.6}&{72.8} 
			&{47.2}&{47.2}&{30.5} 
			&{68.4}&{78.8}&{63.1}
			&{84.2}&{90.1}&{82.2}
			&{49.0}&{52.6}&{36.2}\\	 
			TCC$_1$ \cite{shen2021you}  & \checkmark  &{79.0}&\underline{90.6}&{73.3} 
			&{47.9}&{49.1}&{31.2} 
			&{73.2}&{81.4}&{68.9}
			&{84.8}&{89.7}&{82.5}
			&{55.4}&{59.5}&{41.7}\\	
			SPICE$_1$ \cite{niu2022spice} & \checkmark  &{73.4}&{83.8}&{70.5} 
			&{44.8}&{46.8}&{29.4} 
			&\underline{81.7}&\underline{90.8}&\underline{81.2}
			&{82.8}&{92.1}&{83.6}
			&\underline{57.2}&\underline{64.6}&\underline{47.9}\\		
			SeCu$_1$ \cite{qian2023stable}  & \checkmark &{79.9}&{88.5}&{78.2} 
			&\underline{51.6}&\underline{51.6}&\underline{36.0} 
			&{70.7}&{81.4}&{65.7}
			&{-}&{-}&{-}
			&{-}&{-}&{-}\\	
			Temi$_2$$^*$ \cite{adaloglou2023exploring} & \checkmark  &{82.9}&{90.0}&{80.7} 
			&{59.8}&{57.8}&{42.5} 
			&{93.6}&{96.7}&{93.0}
			&{-}&{-}&{-}
			&{-}&{-}&{-}\\	
			\midrule
			DPAC$_1^{J_{pac}}$(\cref{eq:eqn-13})  &  &\underline{81.2}&{89.0}&\underline{79.1} 
			&{48.3}&{50.2}&{34.4} 
			&{81.8}&{89.7}&{80.0}
			&\underline{90.1}&\underline{96.0}&\underline{91.1}
			&{51.9}&{53.9}&{38.9}\\	
			DPAC$_1^{opa}$(\cref{eq:eqn-12}) &  &\textbf{82.7}&\textbf{90.7}&\textbf{81.2} 
			&\textbf{52.9}&\textbf{51.6}&\textbf{36.2} 
			&\textbf{84.5}&\textbf{92.6}&\textbf{84.7}
			&\textbf{90.8}&\textbf{96.2}&\textbf{91.8}
			&\textbf{60.2}&\textbf{65.5}&\textbf{50.0}\\					  	  
			\midrule
			\multicolumn{15}{l}{With self-labeling fine-tuning (†):}\\
			\midrule
			SCAN$_2$†\cite{van2020scan} & \checkmark &{79.7}&{88.3}&{77.2} 
			&{48.6}&{50.7}&{33.3} 
			&{69.8}&{80.9}&{64.6}
			&{-}&{-}&{-}
			&{-}&{-}&{-}\\
			SPICE$_1$†\cite{niu2022spice}  & \checkmark &\underline{86.5}&{92.6}&{85.2} 
			&{56.7}&{53.8}&{38.7} 
			&\textbf{87.2}&\textbf{93.8}&\textbf{87.0}
			&\underline{90.2}&\underline{95.9}&\underline{91.2}
			&\underline{62.7}&\underline{67.5}&\underline{52.6}\\	
			TCL$_1$†\cite{li2022twin} & \checkmark  &{81.9}&{88.7}&{78.0} 
			&{52.9}&{53.1}&{35.7} 
			&{79.9}&{86.8}&{75.7}
			&{87.5}&{89.5}&{83.7}
			&{62.3}&{64.4}&{51.6}\\
			SeCu$_1$†\cite{qian2023stable}  & \checkmark &{86.1}&\underline{93.0}&\underline{85.7} 
			&\textbf{55.2}&\underline{55.1}&\textbf{39.7} 
			&{73.3}&{83.6}&{69.3}
			&{-}&{-}&{-}
			&{-}&{-}&{-}\\	
			\midrule	
			DPAC$_1^{opa}$† & &\textbf{87.0}&\textbf{93.4}&\textbf{86.6} 
			&\underline{54.2}&\textbf{55.5}&\underline{39.3} 
			&\underline{86.3}&\underline{93.4}&\underline{86.1}
			&\textbf{92.5}&\textbf{97.0}&\textbf{93.5}
			&\textbf{66.7}&\textbf{72.6}&\textbf{59.8}\\
			\bottomrule
		\end{tabular}
	}
\end{table}

\subsubsection{Implementation Details}
ResNet-34 \cite{he2016deep} is used as the backbone network in DPAC to ensure a fair comparison. We employed the architecture of SimCLR \cite{chen2020big} with an MLP clustering classifier as model architecture. DPAC incorporates the image transformation of SimCLR as one view of augmentation and randomly selects four transformations from Rand Augment \cite{cubuk2020randaugment} as another view of augmentation. We maintain a consistent set of hyperparameters ($m=1.03, \tau=0.5$) across all amounts of benchmarks. The model is trained for 1,000 epochs in the pre-training step and 200 epochs in the clustering step. As for self-labeling fine-tuning, we utilize a linear classifier and train the model as \cite{li2022twin}. The thresholds are set to 0.95 for each dataset to select sufficient pseudo labels from clustering classifier outputs. Adam \cite{kingma2014adam} with a constant learning rate of $1\times10^{-4}$ and a weight decay of $1\times10^{-4}$ was employed. The batch size is set as 240 and the experiments are implemented on a single NVIDIA 4090 24G GPU.

\subsubsection{Comparison with State of the Arts}
The comparison of DPAC is presented in \cref{tab:tab4}, where methods with additional cluster size constraints are marked. We have the following observations: (1) DPAC significantly surpasses the performance of SimCLR+PAC in \cref{tab:tab3} across all benchmarks. The accuracy of DPAC exceeds PAC by more than 10\% on CIFAR-100, STL-10, and ImageNet-Dogs benchmarks, which demonstrates the semantic learning ability of DPAC. (2) Compared with DPAC$^{J_{pac}}$, DPAC$^{opa}$ has better performance. We attribute this to the fact that the self-labeling manner of OPA alleviates the intrinsic bias brought by the objective function of feature clustering. (3) Compared with deep clustering methods with offline K-means, such as IDFD \cite{tao2021clustering} and PCL \cite{li2020prototypical}, DPAC has superior performance on all benchmarks due to the stable learning offered by the online manner. (4) Compared with online contrastive clustering methods CC \cite{li2021contrastive}, TCC \cite{shen2021you}, and TCL \cite{li2022twin}, DPAC incorporates global spatial information to achieve a fine-grained partitioning of cluster boundaries. (5) Compared with balanced clustering methods and minimal cluster size constraint SeCU \cite{li2022twin}, DPAC omits clustering regular term, is more concise, and outputs more flexible cluster assignments. (6) DPAC$^{opa}$† demonstrates the remarkable extensibility of our approach, showcasing the potential for integration with diverse deep modules.

\begin{table}[t]
	\centering
	\scriptsize
	\caption{Further analysis for DPAC.}
	\label{tab:tab5}
	\begin{subtable}[h]{0.3\textwidth}
		\centering
		\caption{Comparison of different contrastive framework on CIFAR-10.}
		\label{tab:tab5a}
		\begin{tabular}{l|c}
			\toprule
			Method  &ACC   \\
			\midrule 
			SimCLR+OPA  &89.7\\
			MoCo+OPA  &86.5\\
			\midrule
			DPAC$^{opa}$  &90.8\\
			\bottomrule
		\end{tabular}
	\end{subtable}
	\hfill
	\begin{subtable}[h]{0.3\textwidth}
		\centering
		\caption{Comparison of AE based clustering methods on MNIST.}
		\label{tab:tab5b}
		\begin{tabular}{l|cc}
			\toprule
			Method  &NMI   &ACC\\
			\midrule 
			DEC \cite{xie2016unsupervised}  &86.7&88.1\\
			IDEC \cite{guo2017improved} &86.7&88.1\\
			EDESC \cite{cai2022efficient}  & 86.2 &91.3\\
			SSC \cite{wang2023self}  &95.0&98.2\\
			\midrule
			AE+OPA  &90.3&95.4\\	
			\bottomrule
		\end{tabular}
	\end{subtable}
	\hfill
	\begin{subtable}[h]{0.3\textwidth}
		\centering
		\caption{Effect of clustering regularization (CR) term on CIFAR-10. Metric: ACC (\%).}
		\label{tab:tab5c}
		\begin{tabular}{l|cc}
			\toprule
			Method   &w/ CR & w/o CR\\
			\midrule
			SCAN \cite{van2020scan}  &85.7 & 0.1\\
			CC \cite{li2021contrastive} &79.2 & 68.7\\
			GCC \cite{zhong2021graph}  &85.6 & 68.0\\
			\midrule
			DPAC$^{J_{pac}}$  &88.0 & 89.0\\
			DPAC$^{opa}$   &0.1 & 90.7\\
			\bottomrule
		\end{tabular}
	\end{subtable}
\end{table}

\subsubsection{Contrastive Framework Analysis}
We further analyze our DPAC model from different perspectives. We study the effect of proposed contrastive learning. We replace the weighted contrastive loss in  \cref{eq:eqn-13} with standard contrastive loss, and denote it as SimCLR+OPA. Besides, we also perform OPA based on MoCo \cite{he2020momentum}. Conventional contrastive loss treats corresponding augmented samples as positive pairs and others as negative pairs, which ignores the latent semantic structure between negative pairs, leading to the class collision issue \cite{wang2020understanding}.  \cref{tab:tab5a} illustrates our weighted contrastive loss alleviates the cluster collision problem and encodes cluster knowledge into contrastive representation learning. 

\subsubsection{Pretext Task Analysis}
We study the effect of different pretext tasks combined with DPAC. The autoencoder (AE) is used as architecture to prove the universality of our module. The clustering results on MNIST are shown in \cref{tab:tab5b}, which demonstrates that OPA can combine with other self-supervised approaches. Especially, compared with center-based IDEC \cite{xie2016unsupervised} and SSC \cite{wang2023self}, our OPA does not require K-means to initialize cluster layer and has higher scalability.

\subsubsection{Balanced Constraint Analysis}
We study the impact of balanced constraints in different deep clustering methods. Most existing online deep clustering methods \cite{shen2021you, li2021contrastive, zhong2021graph} introduce an average entropy as clustering regularization (CR) term to balance the cluster distribution. The clustering regularization experiments are shown in  \cref{tab:tab5c}. SCAN classifies all samples into a single cluster, and CC and GCC descend into a suboptimal solution without (w/o) the CR term. Besides, if the CR term is too large in the total loss, it will affect the clustering performance in these methods. It is noteworthy that DPAC avoids crashes without the CR term. 
The performance of DPAC$^{J_{pac}}$ with (w/) CR term becomes worse. It demonstrates the superiority of unconstrained clustering, that is, no trade-off between trivial solutions and performance. And DPAC$^{opa}$ with  CR term yields a uniform distribution with no predictive effect. The reason that the constraint of the CR term is too strong so that the classifier cannot accumulate optimization enough information for OPA.

\subsubsection{Hyperparameter Analysis}
As listed in \cref{alg:alg2}, weight exponent $m$ is the key hyperparameter for OPA, $\alpha=1/(m-1)$ is the power of $\boldsymbol{s}_{i,k}$ that amplifies clustering score in \cref{eq:eqn-9} to become sharper to obtain distinguishable cluster assignments. The larger $m$ becomes, the smaller the sharpening effect is, so the model tends to uniform assignments, and clustering may fail due to insufficient scaling. The performance of OPA with different $m$ settings is evaluated in \cref{tab:tab6}. As features become more and more inseparable, the optimal range of $m$ narrows. Therefore, we suggest setting $m$ close to 1 to obtain a universal hyperparameter setting ($m=1.03$ for all datasets). 

\begin{table}[t]
	\centering
	\scriptsize
	\caption{Hyperparameter analysis of exponent $m$ in OPA.  Metrics:ACC (\%).}
	\label{tab:tab6}
		\begin{tabular}{l|ccccccc}
			\toprule
			Weight exponent $m$  &1.01 &1.04&1.07&1.1&1.13&1.16&1.2 \\
			\midrule
			$\alpha=1/(m-1)$ &100.0 &25.0&14.3&10.0&7.7&6.3&5.0 \\
			\midrule
			CIFAR-10  &90.8 &90.3&90.1&89.3&89.3&89.3&10.0\\
		    STL-10  &92.4 &92.1&92.0&91.8&90.7&10.0&10.0\\
			CIFAR-100   &50.2 &51.1&51.0&5.0&5.0&5.0&5.0\\
			\bottomrule
	\end{tabular}
\end{table}

\subsubsection{Superiority of Online Clustering}
We perform the offline clustering version of DPAC to facilitate a comparative analysis between online and offline clustering strategies. We adopt KM, FCM, and PAC to compute offline codes of all samples for \cref{eq:eqn-10} every 1, 10, and 200 epochs. The performance and training duration are reported in \cref{tab:tab7}. It can be observed that the performance of KM and FCM gradually deteriorates as the update frequency decreases, whereas DPAC$^{opa}$ exhibits superior performance and lower time complexity. 

\begin{figure}[t]
	\centering
	\includegraphics[width=0.7\linewidth]{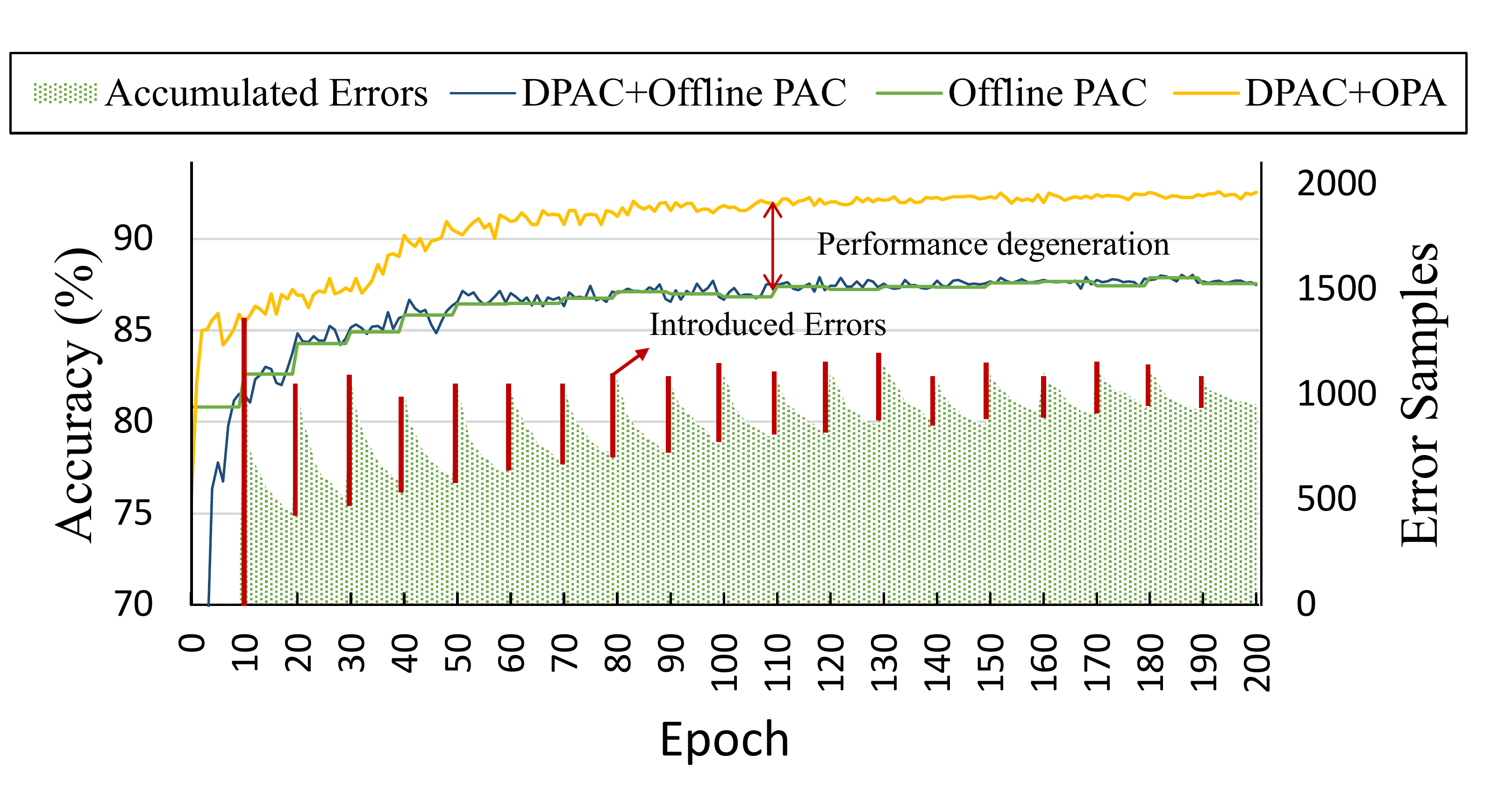}
	\caption{Training process and error accumulation of online and offline DPAC on STL-10.}
	\label{fig:fig2}
\end{figure}

We recorded accumulated errors during DPAC + offline PAC training progress to analyze the error accumulation issue. Offline PAC was conducted every 10 epochs. As depicted in \cref{fig:fig2}, errors (network classifies correctly while offline clustering classifies incorrectly) are introduced by offline clustering every 10 epochs and continue to accumulate through the training process. It demonstrates our OPA module effectively mitigates performance degeneration and error accumulation issues to perform stable and efficient clustering.

\begin{table}[t]
	\centering
	\scriptsize
	\caption{The comparison of online and offline DPAC on STL-10. Metrics: Hour/ACC (\%). }
	\label{tab:tab7}
	\centering
	\begin{tabular}{l|c|c|c}
		\toprule
		\multirow{2.6}{*}{\textbf{Method}} 	 &\multicolumn{3}{c}{Number of Offline Clustering Runs} \\
		\cmidrule(r){2-4}
		&200 &20 &1 \\
		\midrule
		DPAC + offline KM &6.3 / 73.7 &3.0 / 72.0 &{2.0} / 69.3\\
		DPAC + offline FCM &8.5 / 78.7 &3.3 / 77.5 &{2.0} / 68.4\\
		DPAC + offline PAC  &52.2 / 83.7 &6.7 / {87.5} &2.4 / 81.3\\
		\midrule
		DPAC$^{opa}$  &\multicolumn{3}{c}{{2.0} / {92.6}}\\
		\bottomrule
	\end{tabular}
\end{table}

\section{Conclusion}
A novel machine clustering method PAC without cluster center was proposed from a very new perspective, which addresses the shortcomings of center-based clustering approaches and is well-suited for integration with deep models. A theoretical model and an elegant iterative optimization solution for PAC have been developed. PAC implements clustering through sample probability aggregation, which makes part samples based calculation possible. Therefore, an online deep clustering framework DPAC has been developed, which has no constraints on cluster size and can perform more flexible clustering. Experiments on several benchmarks verified the effectiveness of our proposal.
 
\clearpage  

\section*{Acknowledgements}
This work was supported by National Natural Science Foundation of China (Grant No. U22B2036).

%
%
\bibliographystyle{splncs04}
\bibliography{main}
\end{document}